\documentclass[11pt,a4paper]{article}
\usepackage[hyperref]{naaclhlt2019}
\usepackage{times}
\usepackage{latexsym}
\usepackage{url}
\usepackage{tikz,pgfplots}
\usepackage{amsmath}
\usepackage{varwidth}
\usepackage{booktabs}
\usepackage{hyphenat}
\usepackage{amssymb}
\usepackage{tikz-dependency}
\usepackage{multirow}
\usetikzlibrary{patterns,calc,shapes,backgrounds,shadows,positioning,fit,matrix,shapes.geometric}
\tikzset{
smallnode/.style={
  circle,
  inner sep=0pt,
  text width=5mm,
  align=center,
  fill=white
  }
}

\aclfinalcopy 
\def\aclpaperid{1086} 

\title{Structural Neural Encoders for AMR-to-text Generation}

\author{Marco Damonte
  \quad Shay B. Cohen
  \\ \normalsize{School of
    Informatics, University of Edinburgh}\\ \normalsize{10 Crichton Street,
    Edinburgh EH8 9AB, UK}
  \\ \tt{m.damonte@sms.ed.ac.uk} \\
  \tt{scohen@inf.ed.ac.uk}}

\date{}

\begin{document}
\maketitle
\begin{abstract}
AMR-to-text generation is a problem recently introduced to the NLP community, in which the goal is to generate sentences from Abstract Meaning Representation (AMR) graphs. 
Sequence-to-sequence models can be used to this end by converting the AMR graphs to strings.
Approaching the problem while working directly with graphs requires the use of graph-to-sequence models that encode the AMR graph into a vector representation.
Such encoding has been shown to be beneficial in the past, and unlike sequential encoding, it allows us to explicitly capture reentrant structures in the AMR graphs.
We investigate the extent to which reentrancies (nodes with multiple parents) have an impact on AMR-to-text generation by comparing graph encoders to tree encoders, where reentrancies are not preserved. We show that improvements in the treatment of reentrancies and long-range dependencies contribute to higher overall scores for graph encoders.
Our best model achieves 24.40 BLEU on LDC2015E86, outperforming the state of the art by 1.1 points and 24.54 BLEU on LDC2017T10, outperforming the state of the art by 1.24 points.

\end{abstract}

\section{Introduction}

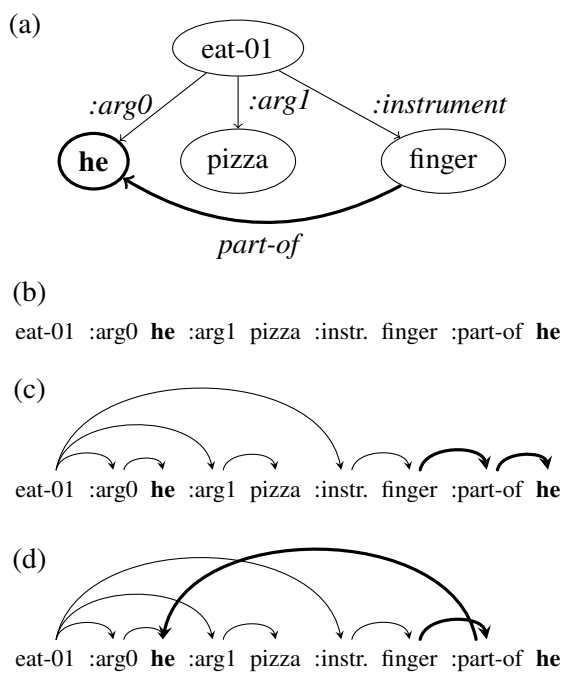
\begin{figure}[ht!]
  \begin{tikzpicture}
    \draw (-3,10.3) node(a) {(a)};
    \draw (-0.2,10) node(beg)[ellipse,draw] {eat-01};
    \draw (-2.1,8.5) node(i)[ellipse,draw, very thick] {\textbf{he}};
    \draw (-0.2,8.5) node(you)[ellipse,draw] {pizza};
    \draw (2.5,8.5) node(excuse)[ellipse,draw] {finger};
    \draw [->] (beg) -- node[left]{\emph{:arg0}} (i);
    \draw [->] (beg) -- node[right]{\emph{:arg1}} (you);
    \draw [->] (beg) -- node[right=0.3cm]{\emph{:instrument}} (excuse);
    \draw[bend left,->, very thick]  (excuse) to node [below] {\emph{part-of}} (i);    
  \end{tikzpicture}
  \begin{dependency}[theme = simple]
   \begin{deptext}
      \small{eat-01} \& \small{:arg0} \& \small{\textbf{he}} \& \small{:arg1} \& \small{pizza} \& \small{:instr.} \& \small{finger} \& \small{:part-of} \& \small{\textbf{he}} \\
   \end{deptext}
   \node (tmp) [below left of = \wordref{1}{1}, xshift = 0.5cm, yshift=1.2cm] {(b)};
\end{dependency} 
  \begin{dependency}[theme = simple]
   \begin{deptext}
      \small{eat-01} \& \small{:arg0} \& \small{\textbf{he}} \& \small{:arg1} \& \small{pizza} \& \small{:instr.} \& \small{finger} \& \small{:part-of} \& \small{\textbf{he}} \\
   \end{deptext}
   \depedge{1}{2}{}
   \depedge{2}{3}{}
   \depedge{1}{4}{}
   \depedge{4}{5}{}
   \depedge{1}{6}{}
   \depedge{6}{7}{}
   \depedge[edge style={very thick}]{7}{8}{}
   \depedge[edge style={very thick}]{8}{9}{}
   \node (tmp) [below left of = \wordref{1}{1}, xshift = 0.5cm, yshift=2cm] {(c)};
\end{dependency}
  \begin{dependency}[theme = simple]
   \begin{deptext}
      \small{eat-01} \& \small{:arg0} \& \small{\textbf{he}} \& \small{:arg1} \& \small{pizza} \& \small{:instr.} \& \small{finger} \& \small{:part-of} \& \small{\textbf{he}} \\
   \end{deptext}
   \depedge{1}{2}{}
   \depedge{2}{3}{}
   \depedge{1}{4}{}
   \depedge{4}{5}{}
   \depedge{1}{6}{}
   \depedge{6}{7}{}
   \depedge[edge style={very thick}]{7}{8}{}
   \depedge[edge style={very thick}]{8}{3}{}
   \node (tmp) [below left of = \wordref{1}{1}, xshift = 0.5cm, yshift=2cm] {(d)};
\end{dependency}
 \caption{(a) AMR for the sentence \emph{He ate the pizza with his fingers} and different input representations: (b) sequential; (c) tree-structured; (d) graph-structured. The nodes and edges in bold highlight a reentrancy.}
 \label{fig:example}
\end{figure}

Abstract Meaning Representation (AMR; \citealt{Banarescu13abstractmeaning}) is a semantic graph representation that abstracts away from the syntactic realization of a sentence, where nodes in the graph represent concepts and edges represent semantic relations between them. AMRs are graphs, rather than trees, because co-references and control structures result in nodes with multiple parents, called reentrancies. For instance, the AMR of Figure~\ref{fig:example}(a) contains a reentrancy between \emph{finger} and \emph{he}, caused by the possessive pronoun \emph{his}. AMR-to-text generation is the task of automatically generating natural language from AMR graphs.

Attentive encoder/decoder architectures, commonly used for Neural Machine Translation (NMT), have been explored for this task \cite{konstas2017neural,song,beck}. 
In order to use sequence-to-sequence models, \newcite{konstas2017neural} reduce the AMR graphs to sequences, while \newcite{song} and \newcite{beck} directly encode them as graphs. Graph encoding allows the model to explicitly encode reentrant structures present in the AMR graphs.
While central to AMR, reentrancies are often hard to treat both in parsing and in generation. Previous work either removed them from the graphs, hence obtaining sequential \cite{konstas2017neural} or tree-structured \cite{liu2018toward,takase2016neural} data, while other work maintained them but did not analyze their impact on performance \cite[e.g.,][]{song,beck}. \newcite{damonte2016incremental} showed that state-of-the-art parsers do not perform well in predicting reentrant structures, while \newcite{van2017dealing} compared different pre- and post-processing techniques to improve the performance of sequence-to-sequence parsers with respect to reentrancies. It is not yet clear whether explicit encoding of reentrancies is beneficial for generation.

In this paper, we compare three types of encoders for AMR: 1) sequential encoders, which reduce AMR graphs to sequences; 2) tree encoders, which ignore reentrancies; and 3) graph encoders. 
We pay particular attention to two phenomena: reentrancies, which mark co-reference and control structures, and long-range dependencies in the AMR graphs, which are expected to benefit from structural encoding. The contributions of the paper are two-fold:
\begin{itemize}
  \item We present structural encoders for the encoder/decoder framework and show the benefits of graph encoders not only compared to sequential encoders but also compared to tree encoders, which have not been studied so far for AMR-to-text generation.
  \item We show that better treatment of reentrancies and long-range dependencies contributes to improvements in the graph encoders.
\end{itemize}

Our best model, based on a graph encoder, achieves state-of-the-art results for both the LDC2015E86 dataset (24.40 on BLEU and 23.79 on Meteor) and the LDC2017T10 dataset (24.54 on BLEU and 24.07 on Meteor).

\section{Input Representations}
\label{sec:input}

\paragraph{Graph-structured AMRs}
AMRs are normally represented as rooted and directed graphs:
\begin{align*}
\begin{split}
& G_0 = (V_0, E_0, L), \\
& V_0 = \{v_1, v_2, \dots, v_n \},\\ 
& root \in V_0,
\end{split}
\end{align*}
\noindent where $V_0$ are the graph vertices (or nodes) and $root$ is a designated root node in $V_0$. The edges in the AMR are labeled:
\begin{align*}
\begin{split}
& E_0 \subseteq V_0 \times L \times V_0, \\
& L = \{\ell_1, \ell_2, \dots, \ell_{n'} \}.
\end{split}
\end{align*}
Each edge $e \in E_0$ is a triple: $e=(i,label,j)$, where $i \in V_0$ is the parent node, $label \in L$ is the edge label and $j \in V_0$ is the child node.

In order to obtain unlabeled edges, thus decreasing the total number of parameters required by the models, we replace each labeled edge $e = (i, label, j)$ with two unlabeled edges: $e_1 = (i, label), e_2 = (label, j)$:
\begin{align*}
\begin{split}
& G = (V, E), \\
& V = V_0 \cup L = \{v_1, \dots, v_n, \ell_1, \dots, \ell_{n'} \},\\
& E \subseteq (V_0 \times L) \cup (L \times V_0).
\end{split}
\end{align*}
Each unlabeled edge $e \in E$ is a pair: $e=(i,j)$,
where one of the following holds:
\begin{enumerate}
\item $i \in V_0$ and $j \in L$;
\item $i \in L$ and $j \in V_0$.
\end{enumerate}

For instance, the edge between \emph{eat-01} and \emph{he} with label \emph{:arg0} of Figure~\ref{fig:example}(a) is replaced by two edges in Figure~\ref{fig:example}(d): an edge between \emph{eat-01} and \emph{:arg0} and another one between \emph{:arg0} and \emph{he}.
The process, also used in \newcite{beck}, tranforms the input graph into its equivalent Levi graph \cite{levi1942finite}. 

\paragraph{Tree-structured AMRs} 
In order to obtain tree structures, it is necessary to discard the reentrancies from the AMR graphs. 
Similarly to \newcite{takase2016neural}, we replace nodes with $n > 1$ incoming edges with $n$ identically labeled nodes, each with a single incoming edge.


\paragraph{Sequential AMRs} Following \newcite{konstas2017neural}, the input sequence is a linearized and anonymized AMR graph. Linearization is used to convert the graph into a sequence:
\begin{align*}
\begin{split}
& x = x_1, \dots, x_N, \\
& x_i \in V.
\end{split}
\end{align*}
The depth-first traversal of the graph defines the indexing between nodes and tokens in the sequence. For instance, the root node is $x_1$, its leftmost child is $x_2$ and so on. 
Nodes with multiple parents are visited more than once. At each visit, their labels are repeated in the sequence, effectively losing reentrancy information, as shown in Figure~\ref{fig:example}(b).

Anonymization removes names and rare words with coarse categories to reduce data sparsity. 
An alternative to anonymization is to employ a copy mechanism \cite{gulcehre2016pointing}, where the models learn to copy rare words from the input itself. In this paper, we follow the anonymization approach. 

\section{Encoders}

In this section, we review the encoders adopted as building blocks for our tree and graph encoders. 

\subsection{Recurrent Neural Network Encoders}
\label{sec:sequential}

We reimplement the encoder of \newcite{konstas2017neural}, where the sequential linearization is the input to a bidirectional LSTM (BiLSTM; \citealt{graves2013speech}) network. The hidden state of the BiLSTM at step $i$ is used as a context-aware word representation of the $i$-th token in the sequence:
\begin{align*}
  e_{1:N}=\mathrm{BiLSTM}(x_{1:N}),
\end{align*}
where $e_i \in \mathbb{R}^{d}$, $d$ is the size of the output embeddings. 

\subsection{TreeLSTM Encoders}
\label{sec:tree_encoders}

Tree-Structured Long Short-Term Memory Networks (TreeLSTM; \citealt{tai2015improved}) have been introduced primarily as a way to encode the hierarchical structure of syntactic trees \cite{tai2015improved}, but they have also been applied to AMR for the task of headline generation \cite{takase2016neural}. 
TreeLSTMs assume tree-structured input, so AMR graphs must be preprocessed to respect this constraint: reentrancies, which play an essential role in AMR, must be removed, thereby transforming the graphs into trees. 

We use the Child-Sum variant introduced by \newcite{tai2015improved}, which processes the tree in a bottom-up pass. When visiting a node, the hidden states of its children are summed up in a single vector which is then passed into recurrent gates.

In order to use information from both incoming and outgoing edges (parents and children), we employ bidirectional TreeLSTMs \cite{eriguchi2016tree}, where the bottom-up pass is followed by a top-down pass. The top-down state of the root node is obtained by feeding the bottom-up state of the root node through a feed-forward layer:
\begin{align*}
h_{\mathrm{root}}^{\downarrow} = \mathrm{tanh} (W_r h_{\mathrm{root}}^{\uparrow} + b),
\end{align*}
where $h_i^{\uparrow}$ is the hidden state of node $x_i \in V$ for the bottom-up pass and $h_i^{\downarrow}$ is the hidden state of node $x_i$ for the top-down pass.

The bottom up states for all other nodes are computed with an LSTM, with the cell state given by their parent nodes:
\begin{align*}
h_{i}^{\downarrow} = \mathrm{LSTM} (h_{p(i)}^{\uparrow}, h_{i}^{\uparrow}),
\end{align*}
where $p(i)$ is the parent of node $x_i$ in the tree.
The final hidden states are obtained by concatenating the states from the bottom-up pass and the top-down pass:
 \begin{align*}
h_{i} = \big[ h_{i}^{\downarrow} ; h_{i}^{\uparrow} \big] .
\end{align*}

The hidden state of the root node is usually used as a representation for the entire tree. In order to use attention over all nodes, as in traditional NMT \cite{bahdanau2014neural}, we can however build node embeddings by extracting the hidden states of each node in the tree:
\begin{align*}
e_{1:N} = h_{1:N}, 
\end{align*}
where $e_i \in \mathbb{R}^{d}$, $d$ is the size of the output embeddings.

The encoder is related to the TreeLSTM encoder of \newcite{takase2016neural}, which however encodes labeled trees and does not use a top-down pass.

\subsection{Graph Convolutional Network Encoders}
\label{sec:graph_encoders}

Graph Convolutional Network (GCN; \citealt{duvenaud2015convolutional,kipf2016semi}) is a neural network architecture that learns embeddings of nodes in a graph by looking at its nearby nodes. 
In Natural Language Processing, GCNs have been used for Semantic Role Labeling \cite{marcheggiani2017encoding}, NMT \cite{bastings2017graph}, Named Entity Recognition \cite{cetoli2017graph} and text generation \cite{diego}.


A graph-to-sequence neural network was first introduced by \newcite{xu2018graph2seq}. The authors review the similarities between their approach, GCN and another approach, based on GRUs \cite{li2015gated}. The latter recently inspired a graph-to-sequence architecture for AMR-to-text generation \cite{beck}. Simultaneously, \newcite{song} proposed a graph encoder based on LSTMs. 

The architectures of \newcite{song} and \newcite{beck} are both based on the same core computation of a GCN, which sums over the embeddings of the immediate neighborhood of each node: 
\begin{align*}
h_i^{(k + 1)} = \sigma \Bigg( \sum_{j \in \mathcal{N}(i)}{W_{(j,i)}^{(k)}} h_j^{(k)} + b^{(k)}  \Bigg),
\end{align*}
where $h_i^{(k)}$ is the embeddings of node $x_i \in V$ at layer $k$, $\sigma$ is a non-linear activation function, $\mathcal{N}(i)$ is the set of the immediate neighbors of $x_i$, $W_{(j,i)}^{(k)} \in \mathbb{R}^{m \times m}$ and $b^{(k)} \in \mathbb{R}^{m}$, with $m$ being the size of the embeddings.
  
It is possible to use recurrent networks to model the update of the node embeddings. Specifically, \newcite{beck} uses a GRU layer where the gates are modeled as GCN layers. 
\newcite{song} did not use the activation function $\sigma$ and perform an LSTM update instead. 

The systems of \newcite{song} and \newcite{beck} further differ in design and implementation decisions such as in the use of edge label and edge directionality. 
Throughout the rest of the paper, we follow the traditional, non-recurrent, implementation of GCN also adopted in other NLP tasks \cite{marcheggiani2017encoding,bastings2017graph,cetoli2017graph}. In our experiments, the node embeddings are computed as follows:
\begin{equation}
h_i^{(k + 1)} = \sigma \Bigg( \sum_{j \in \mathcal{N}(i)}{W_{\mathrm{dir}(j,i)}^{(k)}} h_j^{(k)} + b^{(k)} \Bigg),
\label{eq:gcn_final}
\end{equation}
where $\mathrm{dir}(j,i)$ indicates the direction of the edge between $x_j$ and $x_i$ (i.e., outgoing or incoming edge). 
The hidden vectors from the last layer of the GCN network are finally used to represent each node in the graph:
\begin{align*}
e_{1:N} = h_1^{(K)}, \dots, h_N^{(K)},
\end{align*}
where K is the number of GCN layers used, $e_i \in \mathbb{R}^{d}$, $d$ is the size of the output embeddings. 

To regularize the models we apply dropout \cite{srivastava2014dropout} as well as edge dropout \cite{marcheggiani2017encoding}. We also include highway connections \cite{srivastava2015highway} between GCN layers. 

While GCN can naturally be used to encode graphs, they can also be applied to trees by removing reentrancies from the input graphs. In the experiments of Section~\ref{sec:experiments}, we explore GCN-based models both as graph encoders (reentrancies are maintained) as well as tree encoders (reentrancies are ignored). 

\section{Stacking Encoders}
\label{sec:stack}

\begin{figure}[!t]
 \centering
  \begin{tikzpicture}
    \tiny{
    \draw (4.7,3.7) node(beg)[ellipse,draw] {$x_1$};
    \draw (4,3.2) node(i)[ellipse,draw] {$x_2$};
    \draw (4.7,3.2) node(you)[] {$\dots$};
    \draw (5.4,3.2) node(you)[ellipse,draw] {$x_N$};
    \draw [->] (beg) -- node[left]{} (i);
    \draw [->] (beg) -- node[right]{} (you);
    }
    \draw (4.7,5.5) node(gcn_l) [draw, rectangle, minimum width=2cm, minimum height=0.7cm, fill=red!15, rounded corners=0.1cm] {GCN/TreeLSTM}; 
    \tiny{
      \draw (4.7,7.1) node(rootlup)[ellipse,draw] {$h_1$};
      \draw (4,6.6) node(i)[ellipse,draw] {$h_2$};
      \draw (4.7,6.6) node(dotsldown)[] {$\dots$};
      \draw (5.4,6.6) node(you)[ellipse,draw] {$h_N$};
      \draw [->] (rootlup) -- node[left]{} (i);
      \draw [->] (rootlup) -- node[right]{} (you);
    }
    \draw (4,8) node(hl0) {$h_1$}; \draw (4.5,8) node(hl1) {$h_2$}; \draw (5,8) node(hldots) {$\dots$}; \draw (5.5,8) node(hln) {$h_n$};
    \draw (4.7,8.8) node(bilstm_l) [draw, rectangle, minimum width=2cm, minimum height=0.7cm, fill=yellow!15, rounded corners=0.1cm] {BiLSTM};
    \draw (4,9.7) node(el0) {$e_1$}; \draw (4.5,9.7) node(el1) {$e_2$}; \draw (5,9.7) node(eldots) [minimum height=0.28cm]  {$\dots$}; \draw (5.5,9.7) node(eln) {$e_n$};

    \tiny{
    \draw (1,3.7) node(beg)[ellipse,draw] {$x_1$};
    \draw (0.3,3.2) node(i)[ellipse,draw] {$x_2$};
    \draw (1,3.2) node(you)[] {$\dots$};
    \draw (1.7,3.2) node(you)[ellipse,draw] {$x_N$};
    \draw [->] (beg) -- node[left]{} (i);
    \draw [->] (beg) -- node[right]{} (you);
    }
    \draw (0.3,4.7) node(xr0) {$x_1$}; \draw (0.8,4.7) node(xr1) {$x_2$}; \draw (1.3,4.7) node(xrdots) {$\dots$}; \draw (1.8,4.7) node(xrn) {$x_n$};
    \draw (1,5.5) node(bilstm_r) [draw, rectangle, minimum width=2cm, minimum height=0.7cm, fill=yellow!15, rounded corners=0.1cm] {BiLSTM};
    \draw (0.3,6.4) node(hr0) {$h_1$}; \draw (0.8,6.4) node(hr1) {$h_2$}; \draw (1.3,6.4) node(hrdots) [minimum height=0.33cm] {$\dots$}; \draw (1.8,6.4) node(hrn) {$h_n$};
    \tiny{
    \draw (1,7.7) node(rootrdown)[ellipse,draw] {$h_1$};
    \draw (0.3,7.2) node(i)[ellipse,draw] {$h_2$};
    \draw (1,7.2) node(dotsrdown)[] {$\dots$};
    \draw (1.7,7.2) node(you)[ellipse,draw] {$h_N$};
    \draw [->] (rootrdown) -- node[left]{} (i);
    \draw [->] (rootrdown) -- node[right]{} (you);
    }
    \draw (1,8.8) node(gcn_r) [draw, rectangle, minimum width=2cm, minimum height=0.7cm, fill=red!15, rounded corners=0.1cm] {GCN/TreeLSTM}; 
    \tiny{
      \draw (1,10.3) node(beg)[ellipse,draw] {$e_1$};
      \draw (0.3,9.7) node(i)[ellipse,draw] {$e_2$};
      \draw (1,9.7) node(dotsrup)[] {$\dots$};
      \draw (1.7,9.7) node(you)[ellipse,draw] {$e_N$};
      \draw [->] (beg) -- node[left]{} (i);
      \draw [->] (beg) -- node[right]{} (you);
    }      

    \draw [-, dashed] (3,3.5) -- (3,10);
    \draw [->,>=stealth, thick, dashed] (1,4) -- (1,4.5);
    \draw [->,>=stealth, thick] (xr0.north) -- (bilstm_r.south -| xr0.north);
    \draw [->,>=stealth, thick] (xr1.north) -- (bilstm_r.south -| xr1.north);
    \draw [->,>=stealth, thick] (xrdots.north) -- (bilstm_r.south -| xrdots.north);
    \draw [->,>=stealth, thick] (xrn.north) -- (bilstm_r.south -| xrn.north);

    \draw [->,>=stealth, thick] (bilstm_r.north -| hr0.south) -- (hr0.south);
    \draw [->,>=stealth, thick] (bilstm_r.north -| hr1.south) -- (hr1.south);
    \draw [->,>=stealth, thick] (bilstm_r.north -| hrdots.south) -- (hrdots.south);
    \draw [->,>=stealth, thick] (bilstm_r.north -| hrn.south) -- (hrn.south);    

    \draw [->,>=stealth, thick, dashed] (1,6.6) -- (1,7);
    \draw [->,>=stealth, thick] (1,8) -- (gcn_r.south);
    \draw [->,>=stealth, thick] (gcn_r.north) -- (dotsrup.south);

    \draw [->,>=stealth, thick] (4.7,4) -- (gcn_l.south);
    \draw [->,>=stealth, thick] (gcn_l.north) -- (4.7,6.3);
    \draw [->,>=stealth, thick, dashed] (4.7,7.4) -- (4.7,7.8);
    \draw [->,>=stealth, thick] (hl0.north) -- (bilstm_l.south -| hl0.north);
    \draw [->,>=stealth, thick] (hl1.north) -- (bilstm_l.south -| hl1.north);
    \draw [->,>=stealth, thick] (hldots.north) -- (bilstm_l.south -| hldots.north);
    \draw [->,>=stealth, thick] (hln.north) -- (bilstm_l.south -| hln.north);

    \draw [->,>=stealth, thick] (bilstm_l.north -| el0.south) -- (el0.south);
    \draw [->,>=stealth, thick] (bilstm_l.north -| el1.south) -- (el1.south);
    \draw [->,>=stealth, thick] (bilstm_l.north -| eldots.south) -- (eldots.south);
    \draw [->,>=stealth, thick] (bilstm_l.north -| eln.south) -- (eln.south);  

  \end{tikzpicture}
 \caption{Two ways of stacking recurrent and structural models. Left side: structure on top of sequence, where the structural encoders are applied to the hidden vectors computed by the BiLSTM. Right side: sequence on top of structure, where the structural encoder is used to create better embeddings which are then fed to the BiLSTM. The dotted lines refer to the process of converting the graph into a sequence or vice-versa.}
 \label{fig:arch}
\end{figure}
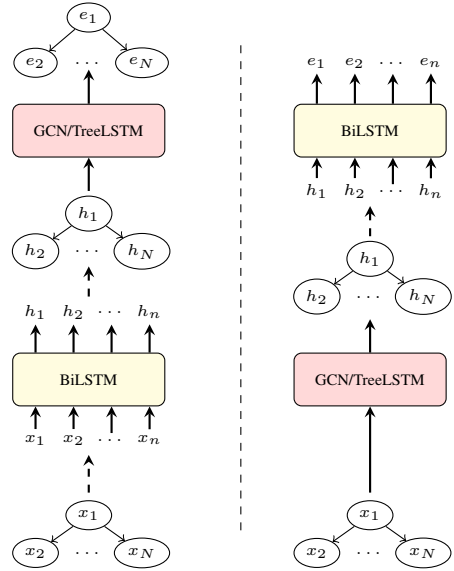

We aimed at stacking the explicit source of structural information provided by TreeLSTMs and GCNs with the sequential information which BiLSTMs extract well. 
This was shown to be effective for other tasks with both TreeLSTMs \cite{eriguchi2016tree,chen2017improved} and GCNs \cite{marcheggiani2017encoding,cetoli2017graph,bastings2017graph}. In previous work, the structural encoders (tree or graph) were used on top of the BiLSTM network: first, the input is passed through the sequential encoder, the output of which is then fed into the structural encoder. While we experiment with this approach, we also propose an alternative solution where the BiLSTM network is used on top of the structural encoder: the input embeddings are refined by exploiting the explicit structural information given by the graph. The refined embeddings are then fed into the BiLSTM networks. See Figure~\ref{fig:arch} for a graphical representation of the two approaches. In our experiments, we found this approach to be more effective. Compared to models that interleave structural and recurrent components such as the systems of \newcite{song} and \newcite{beck}, stacking the components allows us to test for their contributions more easily.

\subsection{Structure on Top of Sequence}
\label{sec:ontopofbilstm}
In this setup, BiLSTMs are used as in Section~\ref{sec:sequential} to encode the linearized and anonymized AMR.
The context provided by the BiLSTM is a sequential one. We then apply either GCN or TreeLSTM on the output of the BiLSTM, by initializing the GCN or TreeLSTM embeddings with the BiLSTM hidden states. We call these models {\sc SeqGCN} and \nohyphens{{\sc SeqTreeLSTM}}.



\subsection{Sequence on Top of Structure}
\label{sec:bilstmontop}

We also propose a different approach for integrating graph information into the encoder, by swapping the order of the BiLSTM and the structural encoder: we aim at using the structured information provided by the AMR graph as a way to refine the original word representations. We first apply the structural encoder to the input graphs. The GCN or TreeLSTM representations are then fed into the BiLSTM. We call these models {\sc GCNSeq} and {\sc TreeLSTMSeq}. 

The motivation behind this approach is that we know that BiLSTMs, given appropriate input embeddings, are very effective at encoding the input sequences. In order to exploit their strength, we do not amend their output but rather provide them with better input embeddings to start with, by explicitly taking the graph relations into account. 

\section{Experiments}
\label{sec:experiments}

We use both BLEU \cite{papineni2002bleu} and Meteor \cite{banerjee2005meteor} as evaluation metrics.\footnote{We used the evaluation script available at \url{https://github.com/sinantie/NeuralAmr}.} We report results on the AMR dataset LDC2015E86 and LDC2017T10.
All systems are implemented in PyTorch \cite{paszke2017automatic} using the framework OpenNMT-py \cite{opennmt}.
Hyperparameters of each model were tuned on the development set of LDC2015E86. For the GCN components, we use two layers, $\mathrm{ReLU}$ activations, and $\mathrm{tanh}$ highway layers. We use single layer LSTMs. We train with SGD with the initial learning rate set to 1 and decay to 0.8. Batch size is set to 100.\footnote{Our code is available at \url{https://github.com/mdtux89/OpenNMT-py-AMR-to-text}.} 

We first evaluate the overall performance of the models, after which we focus on two phenomena that we expect to benefit most from structural encoders: reentrancies and long-range dependencies. Table~\ref{tab:dev_results} shows the comparison on the development split of the LDC2015E86 dataset between sequential, tree and graph encoders. The sequential encoder ({\sc Seq}) is a re-implementation of \newcite{konstas2017neural}. We test both approaches of stacking structural and sequential components: structure on top of sequence ({\sc SeqTreeLSTM} and {\sc SeqGCN}), and sequence on top of structure ({\sc TreeLSTMSeq} and {\sc GCNSeq}). To inspect the effect of the sequential component, we run ablation tests by removing the RNNs altogether ({\sc TreeLSTM} and {\sc GCN}). GCN-based models are used both as tree encoders (reentrancies are removed) and graph encoders (reentrancies are maintained).
\begin{table}
\centering
\begin{tabular}{llcc}
\toprule
\textbf{Input} & \textbf{Model} & \textbf{BLEU} & \textbf{Meteor}\\
\midrule
Seq & {\sc Seq} & 21.40 & 22.00 \\
\midrule
\multirow{6}{*}{Tree} 
 & {\sc SeqTreeLSTM} & 21.84 & 22.34 \\
 & {\sc TreeLSTMSeq} & 22.26 & 22.87 \\
 & {\sc TreeLSTM} & 22.07 & 22.57 \\
 & {\sc SeqGCN} & 21.84 & 22.21 \\
 & {\sc GCNSeq} & \textbf{23.62} & \textbf{23.77} \\
 & {\sc GCN} & 15.83 & 17.76 \\
\midrule
\multirow{3}{*}{Graph} 
 & {\sc SeqGCN} & 22.06 & 22.18 \\
 & {\sc GCNSeq} & \textbf{23.95} & \textbf{24.00} \\
 & {\sc GCN} & 15.94 & 17.76 \\
\bottomrule
\end{tabular}
\caption{BLEU and Meteor (\%) scores on the development split of LDC2015E86.}
\label{tab:dev_results}
\end{table}

For both TreeLSTM-based and GCN-based models, our proposed approach of applying the structural encoder before the RNN achieves better scores. This is especially true for GCN-based models, for which we also note a drastic drop in performance when the RNN is removed, highlighting the importance of a sequential component. On the other hand, RNN layers seem to have less impact on TreeLSTM-based models. This outcome is not unexpected, as TreeLSTMs already use LSTM gates in their computation. 

The results show a clear advantage of tree and graph encoders over the sequential encoder. The best performing model is {\sc GCNSeq}, both as a tree and as a graph encoder, with the latter obtaining the highest results.


\begin{table}
\centering
\begin{tabular}{p{4cm}cc}
\toprule
\textbf{Model} & \textbf{BLEU} & \textbf{Meteor}\\
\midrule    
\multicolumn{3}{c}{LDC2015E86}\\
\midrule
{\sc Seq} & 21.43 & 21.53\\
{\sc Tree} & 23.93 & 23.32\\
{\sc Graph} & \textbf{24.40} & \textbf{23.60}\\
\newcite{konstas2017neural} & 22.00 & - \\
\newcite{song} & 23.30 & - \\
\midrule
\multicolumn{3}{c}{LDC2017T10}\\
\midrule
{\sc Seq} & 22.19 & 22.68 \\
{\sc Tree} & 24.06 & 23.62\\
{\sc Graph} & \textbf{24.54} & \textbf{24.07} \\
\newcite{beck} & 23.30 & - \\
\bottomrule
\end{tabular}
\caption{Scores on the test split of LDC2015E86 and LDC2017T10. {\sc Tree} is the tree-based {\sc GCNSeq} and {\sc Graph} is the graph-based {\sc GCNSeq}.}
\label{tab:test_results}
\end{table}

Table~\ref{tab:test_results} shows the comparison between our best sequential ({\sc Seq}), tree ({\sc GCNSeq} without reentrancies, henceforth called {\sc Tree}) and graph encoders ({\sc GCNSeq} with reentrancies, henceforth called {\sc Graph}) on the test set of LDC2015E86 and LDC2017T10. We also include state-of-the-art results reported on these datasets for sequential encoding \citep{konstas2017neural} and graph encoding \citep{song,beck}.\footnote{We run comparisons on systems without ensembling nor additional data.} 
In order to mitigate the effects of random seeds, we train five models with different random seeds and report the results of the median model, according to their BLEU score on the development set \cite{beck}. We achieve state-of-the-art results with both tree and graph encoders, demonstrating the efficacy of our GCNSeq approach. The graph encoder outperforms the other systems and previous work on both datasets. These results demonstrate the benefit of structural encoders over purely sequential ones as well as the advantage of explicitly including reentrancies. The differences between our graph encoder and that of \newcite{song} and \newcite{beck} were discussed in Section~\ref{sec:graph_encoders}.

\begin{table}
\centering
\begin{tabular}{lcc}
\toprule
\textbf{\# reentrancies} & \textbf{\# dev sents.} & \textbf{\# test sents.}\\
\midrule
0 & 619 & 622 \\
1-5 & 679 & 679 \\
6-20 & 70 & 70 \\
\bottomrule
\end{tabular}
\caption{Counts of reentrancies for the development and test split of LDC2017T10}
\label{tab:stats_reentrancies}
\end{table}

\subsection{Reentrancies}
\label{sec:reentrancies}
\begin{table}
\centering
\begin{tabular}{lccc}
\toprule
\textbf{Model} & \multicolumn{3}{c}{\textbf{Number of reentrancies}}\\
 & 0 & 1-5 & 6-20\\
\midrule
{\sc Seq} & 42.94 & 31.64 & 23.33 \\
{\sc Tree} & +0.63 & +1.41 & +0.76 \\ 
{\sc Graph} & \textbf{+1.67} & \textbf{+1.54} & \textbf{+3.08} \\
\bottomrule
\end{tabular}
\caption{Differences, with respect to the sequential baseline, in the Meteor score of the test split of LDC2017T10 as a function of the number of reentrancies.}
\label{tab:diff_reentr}
\end{table}

Overall scores show an advantage of graph encoder over tree and sequential encoders, but they do not shed light into how this is achieved. Because graph encoders are the only ones to model reentrancies explicitly, we expect them to deal better with these structures. It is, however, possible that the other models are capable of handling these structures implicitly. Moreover, the dataset contains a large number of examples that do not involve any reentrancies, as shown in Table~\ref{tab:stats_reentrancies}, so that the overall scores may not be representative of the ability of models to capture reentrancies.
It is expected that the benefit of the graph models will be more evident for those examples containing more reentrancies. To test this hypothesis, we evaluate the various scenarios as a function of the number of reentrancies in each example, using the Meteor score as a metric.\footnote{For this analysis we use Meteor instead of BLEU because it is a sentence-level metric, unlike BLEU, which is a corpus-level metric.}

Table~\ref{tab:diff_reentr} shows that the gap between the graph encoder and the other encoders is widest for examples with more than six reentrancies. The Meteor score of the graph encoder for these cases is 3.1\% higher than the one for the sequential encoder and 2.3\% higher than the score achieved by the tree encoder, demonstrating that explicitly encoding reentrancies is more beneficial than the overall scores suggest. Interestingly, it can also be observed that the graph model outperforms the tree model also for examples with no reentrancies, where tree and graph structures are identical. This suggests that preserving reentrancies in the training data has other beneficial effects. In Section~\ref{sec:deps} we explore one: better handling of long-range dependencies.


\subsubsection{Manual Inspection}

In order to further explore how the graph model handles reentrancies differently from the other models, we performed a manual inspection of the models' output. We selected examples containing reentrancies, where the graph model performs better than the other models. These are shown in Table~\ref{tab:example_reentrancies}. In Example (1), we note that the graph model is the only one that correctly predicts the phrase \emph{he finds out}. The wrong verb tense is due to the lack of tense information in AMR graphs. In the sequential model, the pronoun is chosen correctly, but the wrong verb is predicted, while in the tree model the pronoun is missing. In Example (2), only the graph model correctly generates the phrase \emph{you tell them}, while none of the models use \emph{people} as the subject of the predicate \emph{can}. In Example (3), both the graph and the sequential models deal well with the control structure caused by the \emph{recommend} predicate. The sequential model, however, overgenerates a wh-clause. Finally, in Example (4) the tree and graph models deal correctly with the possessive pronoun to generate the phrase \emph{tell your ex}, while the sequential model does not. Overall, we note that the graph model produces a more accurate output than sequential and tree models by generating the correct pronouns and mentions when control verbs and co-references are involved. 

\begin{table*}
\centering
\begin{tabular}{lll}
\toprule
(1)
& {\sc REF} & i dont tell him but \textbf{he finds out} ,\\
& {\sc Seq} & i did n't tell him but \textbf{he was out} .\\
& {\sc Tree} & i do n't tell him but \textbf{found out} .\\
& {\sc Graph} & i do n't tell him but \textbf{he found out} .\\
\midrule
(2)
& {\sc REF} & if \textbf{you tell people} they can help you ,\\
& {\sc Seq} & if \textbf{you tell him} , you can help you !\\
& {\sc Tree} & if \textbf{you tell person\_name\_0 you} , you can help you .\\
& {\sc Graph} & if \textbf{you tell them} , you can help you .\\
\midrule
(3)
& {\sc REF} & \textbf{i 'd recommend} you go and see your doctor too .\\
& {\sc Seq} & \textbf{i recommend} you go to see your doctor who is going to see your doctor .\\
& {\sc Tree} & \textbf{you recommend} going to see your doctor too .\\
& {\sc Graph} & \textbf{i recommend} you going to see your doctor too .\\
\midrule
(4) 
& {\sc REF} & (you) \textbf{tell your ex} that all communication needs to go through the lawyer .\\
& {\sc Seq} & (you) \textbf{tell} that all the communication go through lawyer .\\
& {\sc Tree} & (you) \textbf{tell your ex} , tell your ex , the need for all the communication .\\
& {\sc Graph} & (you) \textbf{tell your ex} the need to go through a lawyer .\\
\bottomrule
\end{tabular}
\caption{Examples of generation from AMR graphs containing reentrancies. {\sc REF} is the reference sentence.}
\label{tab:example_reentrancies}
\end{table*}

\subsubsection{Contrastive Pairs}

For a quantitative analysis of how the different models handle pronouns, we use a method to inspect NMT output for specific linguistic analysis based on contrastive pairs \cite{sennrich2017grammatical}. Given a reference output sentence, a contrastive sentence is generated by introducing a mistake related to the phenomenon we are interested in evaluating. The probability that the model assigns to the reference sentence is then compared to that of the contrastive sentence. The accuracy of a model is determined by the percentage of examples in which the reference sentence has a higher probability than the contrastive sentence. 

We produce contrastive examples by running CoreNLP \cite{corenlp} to identify co-references, which are the primary cause of reentrancies, and introducing a mistake. 
When an expression has multiple mentions, the antecedent is repeated in the linearized AMR. For instance, the linearization of Figure~\ref{fig:example}(b) contains the token \emph{he} twice, which instead appears only once in the sentence. This repetition may result in generating the token \emph{he} twice, rather than using a pronoun to refer back to it. To investigate this possible mistake, we replace one of the mentions with the antecedent (e.g., \emph{John ate the pizza with his fingers} is replaced with \emph{John ate the pizza with John fingers}, which is ungrammatical and as such should be less likely). 

An alternative hypothesis is that even when the generation system correctly decides to predict a pronoun, it selects the wrong one. To test for this, we produce contrastive examples where a pronoun is replaced by either a different type of pronoun (e.g., \emph{John ate the pizza with his fingers} is replaced with \emph{John ate the pizza with him fingers}) or by the same type of pronoun but for a different number (\emph{John ate the pizza with their fingers}) or different gender (\emph{John ate the pizza with her fingers}). Note from Figure~\ref{fig:example} that the graph-structured AMR is the one that more directly captures the relation between \emph{finger} and \emph{he}, and as such it is expected to deal better with this type of mistakes.

From the test split of LDC2017T10, we generated 251 contrastive examples due to antecedent replacements, 912 due to pronoun type replacements, 1840 due to number replacements and 95 due to gender replacements.\footnote{The generated contrastive examples are available at \url{https://github.com/mdtux89/OpenNMT-py}.}
The results are shown in Table~\ref{tab:contrastive}. The sequential encoder performs surprisingly well at this task, with better or on par performance with respect to the tree encoder. The graph encoder outperforms the sequential encoder only for pronoun number and gender replacements. 
Future work is required to more precisely analyze if the different models cope with pronomial mentions in significantly different ways. Other approaches to inspect phenomena of co-reference and control verbs can also be explored, for instance by devising specific training objectives \cite{linzen2016assessing}.

\begin{table}
\centering
\begin{tabular}{lcccc}
\toprule
\textbf{Model} & \textbf{Antec.} & \textbf{Type} & \textbf{Num.} & \textbf{Gender}\\
\midrule
{\sc Seq} & 96.02 & 97.70 & 94.89 & 94.74\\
{\sc Tree} & 96.02 & 96.38 & 93.70 & 92.63\\ 
{\sc Graph} & 96.02 & 96.49 & 95.11 & 95.79\\
\bottomrule
\end{tabular}
\caption{Accuracy (\%) of models, on the test split of LDC201T10, for different categories of contrastive errors: antecedent (Antec.), pronoun type (Type), number (Num.), and gender (Gender).}
\label{tab:contrastive}
\end{table}

\subsection{Long-range Dependencies}
\label{sec:deps}
\begin{table}
\centering
\begin{tabular}{lcc}
\toprule
\textbf{\# max length} & \textbf{\# dev sents.} & \textbf{\# test sents.}\\
\midrule
0-10 & 292 & 307\\
11-50 & 350 & 297\\
51-250 & 21 & 18\\
\bottomrule
\end{tabular}
\caption{Counts of longest dependencies for the development and test split of LDC2017T10}
\label{tab:stats_dependencies}
\end{table}
\begin{table}
\centering
\begin{tabular}{lccc}
\toprule
\textbf{Model} & \multicolumn{3}{c}{\textbf{Max dependency length}}\\
 & 0-10 & 11-50 & 51-200\\
\midrule
{\sc Seq} & 50.49 & 36.28 & 24.14 \\
{\sc Tree} & -0.48 & +1.66 & +2.37 \\ 
{\sc Graph} & \textbf{+1.22} & \textbf{+2.05} & \textbf{+3.04} \\
\bottomrule
\end{tabular}
\caption{Differences, with respect to the sequential baseline, in the Meteor score of the test split of LDC2017T10 as a function of the maximum dependency length.}
\label{tab:diff_deps}
\end{table}

When we encode a long sequence, interactions between items that appear distant from each other in the sequence are difficult to capture. The problem of long-range dependencies in natural language is well known for RNN architectures \cite{bengio1994learning}. Indeed, the need to solve this problem motivated the introduction of LSTM models, which are known to model long-range dependencies better than traditional RNNs.

Because the nodes in the graphs are not aligned with words in the sentence, AMR has no notion of distance between the nodes taking part in an edge. In order to define the length of an AMR edge, we resort to the AMR linearization discussed in Section~\ref{sec:input}. Given the linearization of the AMR $x_1, \dots, x_N$, as discussed in Section~\ref{sec:input}, and an edge between two nodes $x_i$ and $x_j$, the length of the edge is defined as $\vert j - i \vert$. 
For instance, in the AMR of Figure~\ref{fig:example}, the edge between \emph{eat-01} and \emph{:instrument} is a dependency of length five, because of the distance between the two words in the linearization \emph{eat-01 :arg0 he :arg1 pizza :instrument}. 
We then compute the maximum dependency length for each AMR graph.

To verify the hypothesis that long-range dependencies contribute to the improvements of graph models, we compare the models as a function of the maximum dependency length in each example. Longer dependencies are sometimes caused by reentrancies, as in the dependency between \emph{:part-of} and \emph{he} in Figure~\ref{fig:example}. To verify that the contribution in terms of longer dependencies is complementary to that of reentrancies, we exclude sentences with reentrancies from this analysis. Table~\ref{tab:stats_dependencies} shows the statistics for this measure. Results are shown in Table~\ref{tab:diff_deps}. The graph encoder always outperforms both the sequential and the tree encoder. The gap with the sequential encoder increases for longer dependencies. This indicates that longer dependencies are an important factor in improving results for both tree and graph encoders, especially for the latter.


\section{Conclusions}
We introduced models for AMR-to-text generation with the purpose of investigating the difference between sequential, tree and graph encoders. We showed that encoding reentrancies improves overall performance. We observed bigger benefits when the input AMR graphs have a larger number of reentrant structures and longer dependencies. Our best graph encoder, which consists of a GCN wired to a BiLSTM network, improves over the state of the art on all tested datasets. 
We inspected the differences between the models, especially in terms of co-references and control structures. 
Further exploration of graph encoders is left to future work, which may result crucial to improve performance further.

\section*{Acknowledgments}

The authors would like to thank the three anonymous reviewers and Adam Lopez, Ioannis Konstas, Diego Marcheggiani, Sorcha Gilroy, Sameer Bansal, Ida Szubert and Clara Vania for their help and comments. This research was supported by a grant from Bloomberg and by the H2020 project SUMMA, under grant agreement 688139.

\bibliography{naaclhlt2019}
\bibliographystyle{acl_natbib}

\end{document}